\documentclass{article}
\usepackage{amsmath}
\usepackage{amsmath}
\usepackage{amssymb}
\usepackage{algorithm}
\usepackage{algpseudocode}
\usepackage{graphicx}
\usepackage{enumitem}
\usepackage{booktabs}
\usepackage{hyperref}
\usepackage{multicol}

\title{Feedforward Ordering in Neural Connectomes via Feedback Arc Minimization}

\author{
    Soroush Vahidi\thanks{Email: \texttt{sv96@njit.edu}} \\
    New Jersey Institute of Technology, Newark, New Jersey, USA
}
\date{}
\newcommand{\Input}{\State \textbf{Input:} }
\begin{document}
\maketitle
\begin{abstract}
We present a suite of scalable algorithms for minimizing feedback arcs in large-scale weighted directed graphs, with the goal of revealing biologically meaningful feedforward structure in neural connectomes. Using the FlyWire Connectome Challenge dataset, we demonstrate the effectiveness of our ranking strategies in maximizing the total weight of forward-pointing edges. Our methods integrate greedy heuristics, gain-aware local refinements, and global structural analysis based on strongly connected components. Experiments show that our best solution improves the forward edge weight over previous top-performing methods. All algorithms are implemented efficiently in Python and validated using cloud-based execution on Google Colab Pro+.
\end{abstract}
\section{Introduction}
\subsection{Problem Statement}

Let \(G = (V, E, w)\) be a weighted, directed graph representing a neuronal connectome, where \(V\) is the set of \(n\) neurons, \(E\subseteq V \times V\) is the set of synaptic connections, and 
\[
w \colon E \;\to\; \mathbb{R}_{\ge 0}
\]
assigns a nonnegative weight to each directed edge \((u,v)\).  A \emph{permutation} \(\pi\) is a bijection 
\[
\pi \colon V \;\to\; \{1,2,\dots,n\},
\]
which induces a total ordering of the neurons.  An edge \((u,v)\in E\) is called a \emph{forward edge} if \(\pi(u) < \pi(v)\), and a \emph{feedback edge} (or \emph{backward edge}) if \(\pi(u) > \pi(v)\).

Our goal is to find a permutation \(\pi\) that maximizes the total weight of forward edges (equivalently, minimizes the total weight of feedback edges).  Equivalently, we wish to solve
\[
\max_{\pi} \; \sum_{\substack{(u,v)\in E \\ \pi(u)<\pi(v)}} w(u,v),
\]
or, equivalently,
\[
\min_{\pi} \; \sum_{\substack{(u,v)\in E \\ \pi(u)>\pi(v)}} w(u,v).
\]
Finding an exact minimizer for the total weight of feedback edges is known to be NP‐hard \cite{K72} for general graphs of this size, so we seek efficient heuristics and approximation strategies that scale to connectomes with tens of thousands of neurons.
\subsection{Motivation}

A central goal in connectomics is to uncover a meaningful hierarchical ordering of neurons so that most synaptic weight points “forward” (i.e.\ from lower‐ to higher‐level processing) rather than being tied up in recurrent loops.  Recent work has shown that the Drosophila connectome is organized into nested communities at multiple scales, from large‐scale neuropils down to tens of cells \cite{K23}.  However, knowing which modules exist is not enough—one must also determine the dominant direction of information flow between them.  Semedo et al.\ \cite{S22} demonstrated in primate cortex that feedforward and feedback signals occupy distinct population subspaces, so extracting a permutation that minimizes feedback arcs is key to isolating true bottom‐up processing streams without contamination from recurrent inputs.  The idea of finding a minimum‐feedback‐arc set provides exactly this framework: by ordering nodes so as to eliminate as many backward edges as possible, we recover an intrinsic feedforward scaffold in any directed network \cite{Z16}.  From a computational standpoint, Lynn and Bassett \cite{L19} argue that such structural hierarchies (and the corresponding flow constraints) are essential for understanding how network topology shapes dynamics and controllability in real brains.  Moreover, Ramaswamy and Banerjee \cite{R14} show that even without detailed physiology, the wiring alone imposes strict limits on what computations a network can support—so revealing that wiring’s hidden feedforward hierarchy is a prerequisite for any downstream modeling of function or behavior.

\section{Related Work}
Even et al.\ propose approximation algorithms for the weighted feedback vertex set and weighted feedback arc set problems in directed graphs, which directly relate to our goal of minimizing backward edges in neural connectome orderings \cite{E98}. Demetrescu and Finocchi \cite{D03} then present simple combinatorial approximation schemes for minimum feedback arc and vertex set in directed graphs, achieving a bound proportional to the length of the longest simple cycle; their local‐ratio method runs in $O(mn)$ time and works well on dense instances. Schiavinotto and Stützle \cite{S04} provide a detailed search‐space analysis for benchmark instances of the NP‐hard linear ordering problem—examining matrix statistics, autocorrelation, and fitness‐distance correlations—which demonstrates that a memetic algorithm consistently surpasses iterated local search on those benchmarks. Fomin et al.\ \cite{F10} introduce a subexponential‐time local search for weighted feedback arc set in tournaments, efficiently exploring k‐exchange neighborhoods; although tailored to dense tournaments, their use of local refinements inspires our own extended refinement on general graphs. Ostovari and Zarei \cite{O24} improve LP rounding for feedback arc set in weighted tournaments by developing a refined probabilistic rounding function; they achieve a 2.127‐approximation, thus tightening the integrality gap and informing our randomized ranking approach. Vahidi and Koutis \cite{V25} show that a learning‐free, local‐ratio heuristic for MWFAS runs in under a second on large pairwise comparison datasets and consistently yields lower weighted‐upset losses—often achieving “simple loss” below 1.0—outperforming recent learning‐based methods on standard benchmarks. 

In the connectome context, Fornito et al.\ \cite{F13} survey graph‐theoretic approaches to human connectome analysis, emphasizing both methodological advances and pitfalls in modeling structural and functional brain networks; their emphasis on careful network modeling guided our choice to focus on edge‐weighted reorderings. Kuo et al.\ \cite{K16} analyze information flow through layers in feedforward deep neural networks using information‐theoretic measures; their conceptual framing of directed, layered processing motivates our objective of extracting a biologically interpretable feedforward order from the weighted connectome. Borst and Hoffman \cite{B24} formulate connectome seriation as a continuous relaxation by mapping discrete neuron indices to smooth real values and applying gradient descent to minimize feedback arcs; this $O(N^2)$‐time method outperforms classic discrete heuristics on both randomized and biological connectivity matrices. Finally, Khoussainov et al.\ \cite{Z24} develop scalable reduction rules and a divide‐and‐conquer heuristic to find small feedback arc sets in large directed graphs; their preprocessor reduces input size by over 80\% on sparse benchmarks, and the subsequent heuristic improves solution quality by up to 40\% on circuit instances while running in near‐linear time.

\section{Methods}
Our method has a base part that provides an initial order and uses several heuristics to improve the initial order. Algorithm ~\ref{alg:outin} provides the initial order, in which the ratio of sum of the weight of the forward edges to the sum of the weight of all edges is 0.7524. Then, we use Algorithm ~\ref{alg:refine}, Algorithm ~\ref{alg:scc} and Algorithm ~\ref{alg:flat} to improve it to 0.8461.
\subsection{Greedy Ranking via Adaptive Out-Degree Over In-Degree Heuristic}
\label{sec:adaptive_out_inplus1}

This section presents a scalable and interpretable greedy algorithm that ranks nodes based on the adaptive ratio of outgoing to incoming edge weights. The method uses a dynamic out-degree over in-degree heuristic with additive smoothing to avoid division by zero. The approach is iterative and local, adapting node scores as neighbors are removed from the unranked pool.

\subsubsection{Overview}

Given a directed graph $G = (V, E)$ with edge weights $w: E \rightarrow \mathbb{R}_{\ge 0}$, the algorithm assigns each node $u \in V$ an adaptive score:

\begin{equation}
\text{score}(u) = \frac{\text{out}_w(u) + 1}{\text{in}_w(u) + 1}
\end{equation}
where \$\text{out}\_w(u)\$ and \$\text{in}\_w(u)\$ are the total weights of edges going out from and into node \$u\$, respectively. Nodes are ranked greedily based on this score, and scores are updated as neighbors are removed.

\subsubsection{Adaptive Greedy Ranking Procedure}
\begin{algorithm}[H]
\caption{AdaptiveOutOverInPlus1Ranking}\label{alg:outin}
\begin{algorithmic}[1]
\State \textbf{Input:} Edge list $E$ with weights $w$, output file path $f$
\State Initialize $\text{out}_w(u)$ and $\text{in}_w(u)$ for all nodes
\State Compute initial scores: $\text{score}(u) \gets (\text{out}_w(u)+1) / (\text{in}_w(u)+1)$
\State Insert $(-\text{score}(u), u)$ into a max-heap $H$ for all nodes $u$
\State Initialize $\texttt{unranked} \gets V$, $\texttt{ranking} \gets$ empty list
\While{$H$ not empty}
\State Pop $u$ with highest score from $H$
\If{$u \notin \texttt{unranked}$}
\State \textbf{continue}
\EndIf
\State Append $u$ to $\texttt{ranking}$; remove $u$ from $\texttt{unranked}$
\For{each $(u \rightarrow v, w)$}
\If{$v \in \texttt{unranked}$}
\State Update $\text{in}_w(v) \mathrel{-}= w$; recompute $\text{score}(v)$
\State Reinsert $(-\text{score}(v), v)$ into $H$
\EndIf
\EndFor
\For{each $(v \rightarrow u, w)$}
\If{$v \in \texttt{unranked}$}
\State Update $\text{out}_w(v) \mathrel{-}= w$; recompute $\text{score}(v)$
\State Reinsert $(-\text{score}(v), v)$ into $H$
\EndIf
\EndFor
\EndWhile
\If{any unranked nodes remain}
\State Shuffle and append them randomly to $\texttt{ranking}$
\EndIf
\State Write $\texttt{ranking}$ to file $f$
\end{algorithmic}
\end{algorithm}

\subsubsection{Discussion}

This algorithm provides a fast and interpretable baseline for node ranking in large directed graphs. By prioritizing nodes with a higher out-degree relative to their in-degree, the heuristic naturally pushes source-like nodes earlier in the order. The additive smoothing (+1 in both numerator and denominator) ensures numerical stability even for nodes with no in-edges or out-edges.

The adaptive nature of the algorithm—recomputing scores as nodes are removed—introduces a cascading effect that propagates structural information through the graph. This makes the method sensitive to local topology and more robust than static scoring techniques.

Although the algorithm does not guarantee a globally optimal ordering, it performs well in practice and often surpasses simple baselines. Its speed and simplicity make it particularly suitable for massive graphs such as connectomes or web networks.

To ensure completeness, any disconnected or unranked nodes are appended randomly at the end of the ranking. The final ordering can be evaluated using the forward edge ratio:
\begin{equation}
\text{Forward Ratio} \;=\; 
\frac{
  \sum_{\substack{(u,v)\in E \\ \pi(u)<\pi(v)}} w(u,v)
}{
  \sum_{(u,v)\in E} w(u,v)
}
\end{equation}

Using this adaptive greedy algorithm on the FlyWire connectome graph, we obtain a vertex ordering with a forward edge ratio of $0.7524$, demonstrating the practical effectiveness of the method even without global optimization.

\subsection{Extended Refinement with Gain-Aware Backward Edge Correction}
\label{sec:extended_refinement}

In this section, we present an extended local refinement algorithm that iteratively improves a node ranking in a directed graph by converting high-weight \emph{backward edges} into \emph{forward edges}. The goal is to maximize the total weight of forward edges in the final ordering.

\subsubsection{Motivation}

Given a directed graph $G = (V, E)$ with edge weights $w: E \rightarrow \mathbb{R}_{\ge 0}$ and an initial ranking $\pi: V \rightarrow \mathbb{Z}$, a backward edge is one where $\pi(u) > \pi(v)$. The algorithm maintains a max-heap of all current backward edges, ordered by weight. At each step, it selects the heaviest backward edge $(u \rightarrow v)$ and tries to transform the local order around $u$ and $v$ to increase the forward edge weight.

\subsubsection{Extended Reordering Strategy}

Let the nodes ranked strictly between $v$ and $u$ be $n_1, n_2, \dots, n_t$, so the local block is:
\[
[v, n_1, n_2, \dots, n_t, u]
\]
The algorithm searches for an index $r$ such that reordering the block as:
\[
[n_1, \dots, n_r, u, v, n_{r+1}, \dots, n_t]
\]
maximizes the gain in forward edge weight. Only the edges with both endpoints inside this block may change direction, so the gain is computed locally using prefix sums:
\begin{equation}
\Delta \;=\; w_{uv} - w_{vu} \;+\; \mathrm{in}_{v} - \mathrm{out}_{v} \;-\; \mathrm{in}_{u} + \mathrm{out}_{u}
\end{equation}

If no reordering improves the forward weight, a fallback greedy strategy is applied using three candidate transformations: (1) swap $u$ and $v$, (2) push $v$ forward and shift intermediate nodes backward, and (3) pull $u$ backward and shift intermediate nodes forward.

\subsubsection{Pseudocode}

\begin{algorithm}[H]
\caption{RefineRankingWithExtendedStrategy}\label{alg:refine}
\begin{algorithmic}[1]
\State \textbf{Input:} Graph $G = (V, E)$ with weights $w$, initial ranking $\pi$
\State \mbox{Insert all backward edges $(u,v)$ with $\pi(u) > \pi(v)$ into max-heap $H$}
\While{$H$ is not empty}
    \State Pop heaviest edge $(u, v)$ from $H$
    \If{$\pi(u) < \pi(v)$}
        \State \textbf{continue}
    \EndIf
    \State Let $B = \{x \in V \mid \pi(v) < \pi(x) < \pi(u)\}$
    \State Initialize $\texttt{best\_delta} \gets -\infty$
    \For{each index $r = 0$ to $|B|$}
        \State Reorder block to $[n_1, \dots, n_r, u, v, n_{r+1}, \dots]$
        \State Compute local $\Delta$ gain using prefix sums
        \If{$\Delta > \texttt{best\_delta}$}
            \State Update best reordering and best\_delta
        \EndIf
    \EndFor
    \If{$\texttt{best\_delta} > 0$}
        \State Apply best reordering
        \State Update $\pi$ and forward weight
    \Else
        \State Compute greedy gains: swap $(u,v)$, push $v$ forward, or pull $u$ backward
        \If{any greedy gain $> 0$}
            \State Apply best greedy move
            \State Update $\pi$ and forward weight
        \Else
            \State Mark edge as rejected
        \EndIf
    \EndIf
    \State Insert new backward edges involving $u$ or $v$ into $H$
\EndWhile

\If{\textnormal{final ranking improves previous best}}
    \State Save updated ranking to file
\Else
    \State Restore previous best ranking
\EndIf
\end{algorithmic}
\end{algorithm}

\subsection{Topological Refinement within the Largest Strongly Connected Component}
\label{sec:scc_block_refinement}

We present a block-wise refinement algorithm that improves a node ranking by operating only on the largest strongly connected component (SCC) of the input graph. The algorithm increases the total weight of forward edges through a hybrid strategy that combines SCC decomposition, topological sorting, and permutation search within local blocks.

\subsubsection{Overview}

Given a directed graph $G = (V, E)$ with edge weights $w: E \rightarrow \mathbb{R}_{\ge 0}$ and an initial ranking $\pi: V \rightarrow \mathbb{Z}$, the goal is to maximize the total forward weight:
\begin{equation}
\mathrm{FW}(\pi) \;=\; 
\sum_{\substack{(u,v)\in E \\ \pi(u)<\pi(v)}} 
w(u,v)
\end{equation}

The key idea is to focus on the largest SCC of $G$, which typically contains the majority of cyclic structure in real-world graphs. The nodes in this SCC are processed in blocks of fixed size. For each block:
\begin{itemize}
    \item A subgraph is extracted and decomposed into smaller SCCs.
    \item The SCCs are topologically sorted to remove cycles.
    \item Small SCCs (size $\leq 9$) are exhaustively permuted to find the ordering with maximum internal forward weight.
    \item Large SCCs are left in their original rank order.
    \item The reordered block is adopted only if it improves the total forward weight.
\end{itemize}

\subsubsection{Pseudocode}

\begin{algorithm}[H]
\caption{RefineSCCBlocks} \label{alg:scc}
\begin{algorithmic}[1]
\Input Graph $G = (V, E)$ with weights $w$, initial ranking $\pi$, block size $s$
\State Identify largest strongly connected component $C \subseteq V$
\State Let $L \gets$ list of nodes in $C$ sorted by $\pi$
\State Let $B_1, B_2, \dots, B_k$ be consecutive blocks of size $s$ from $L$ (with optional bias)
\For{each block $B$}
    \State Let $G_B \gets G[B]$ be the subgraph induced by $B$
    \State Compute SCCs of $G_B$: $\texttt{sub\_sccs} = \{S_1, \dots, S_m\}$
    \State Build DAG $D$ of SCCs: add edge $S_i \rightarrow S_j$ if some $(u,v) \in E$ with $u \in S_i$, $v \in S_j$
    \State Compute topological order $\sigma$ of $D$
    \State Initialize $O_B \gets [\,]$ (new block order)
    \For{each $S_i$ in $\sigma$}
        \If{$|S_i| \leq 9$}
            \State Try all permutations of $S_i$, select one with max forward weight in $G_B$
            \State Append best permutation to $O_B$
        \Else
            \State Sort $S_i$ by original $\pi$, append to $O_B$
        \EndIf
    \EndFor
    \If{total forward weight with $O_B$ is higher than existing block order}
        \State Update $\pi$ for all nodes in $B$ using $O_B$
        \State Save new ranking if overall forward weight improves
    \Else
        \State Revert to previous $\pi$
    \EndIf
\EndFor
\end{algorithmic}
\end{algorithm}

\subsubsection{Discussion}

This method focuses computational effort where cycles are most prevalent—inside the largest SCC. By working in blocks, it avoids full-graph permutations and enables local improvements that accumulate across steps. Exhaustive search within small SCCs ensures optimal rearrangements, while topological sorting handles larger SCCs efficiently.

The algorithm guarantees that each accepted change improves or maintains the forward weight and is particularly effective on dense or cyclic networks such as biological, transport, or connectome graphs.

\subsection{Flat Partition-Based Block Reordering}
\label{sec:flat_partition_refinement}

In this section, we introduce a score-guided partitioning strategy that incrementally improves node rankings by reordering blocks of nodes based on aggregated edge weights between groups. The goal remains to maximize the total weight of forward edges in the final node ordering.

\subsubsection{Motivation}

Unlike the SCC-based or edge-driven strategies, this method operates by hierarchically partitioning a specified rank interval into balanced groups, evaluating inter-group edge weights, and reordering those groups to increase forward edge weight. The method can be implemented either recursively or non-recursively; in practice, a flat (non-recursive) implementation is often preferred due to its reduced memory usage on large graphs.

\subsubsection{Partitioning Strategy}

Let $\pi: V \rightarrow \mathbb{Z}$ be the current ranking and $w: E \rightarrow \mathbb{R}_{\ge 0}$ the edge weight function. For a given interval of ranks $[s, e]$, the set of nodes whose rank lies in this interval is extracted and sorted. These nodes are divided into $x^\ell$ groups for a fixed arity $x$ and level $\ell$:
\[
P_1, P_2, \dots, P_{x^\ell}
\]
Each group contains approximately equal-sized subsets of nodes with similar scores. These groups are then considered in consecutive windows of size $x$, and we attempt to reorder each window to maximize the total weight of forward edges between parts.

\subsubsection{Evaluation and Reordering}

For each group of $x$ consecutive partitions, all $x!$ permutations are evaluated. Let $W_{i,j}$ denote the total edge weight from group $P_i$ to $P_j$. Then, the forward weight of a given group permutation $\sigma = (g_1, \dots, g_x)$ is:
\[
\text{FW}(\sigma) = \sum_{i < j} W_{g_i, g_j}
\]
The permutation with the highest forward weight is selected. If it differs from the initial order, the nodes in the corresponding partitions are reordered accordingly using their current score values to maintain consistency.

To avoid global reordering side-effects, the new scores are updated incrementally starting from the smallest score in the modified block. This strategy ensures that each permutation refinement remains localized to its group and does not affect scores outside the selected interval.

\subsubsection{Pseudocode}

\begin{algorithm}[H]
\caption{FlatPartitionReorder} \label{alg:flat}
\begin{algorithmic}[1]
\Input Scores $\pi$, indexed edges $E$, arity $x$, level $\ell$, rank interval $[s,e]$
\State Let $S \gets \{v \in V \mid s \leq \pi(v) \leq e\}$
\State Sort $S$ by $\pi$, and partition it into $x^\ell$ groups $P_1, \dots, P_{x^\ell}$ of roughly equal size
\State Build a map from each node to its group index
\For{each $x$-group window $(P_{i_1}, \dots, P_{i_x})$}
    \State Compute edge weights $W_{i,j}$ from $P_i$ to $P_j$ using $E$
    \For{each permutation $\sigma$ of the $x$ groups}
        \State Compute forward weight $\text{FW}(\sigma)$ from $W$
    \EndFor
    \State Let $\sigma^*$ be the permutation with highest forward weight
    \If{$\sigma^*$ differs from current order}
        \State Sort nodes within each $P_i$ by $\pi$
        \State Concatenate reordered groups according to $\sigma^*$
        \State Assign new ranks to reordered nodes using base score $b$
    \EndIf
\EndFor
\State Save updated scores if overall forward weight improves
\end{algorithmic}
\end{algorithm}

\subsubsection{Discussion}

This algorithm provides a scalable and tunable mechanism for refining node rankings. Both recursive and flat implementations are possible; however, recursive versions may consume significantly more memory on large instances. By operating on small blocks and evaluating local improvements, the method efficiently accumulates global gains without requiring expensive full-graph permutations.

\subsection{Global Ranking by Topological Ordering of Strongly Connected Components}
\label{sec:scc_global_ordering}
This section describes a global reordering algorithm that constructs a new ranking by computing a topological order over the strongly connected components (SCCs) of the graph. Unlike local refinement strategies, this method leverages the structure of the graph’s SCCs to form an acyclic condensation, then ranks the nodes to optimize the forward edge weight.

\subsubsection{Overview}
Given a directed graph $G = (V, E)$ with non-negative edge weights $w: E \rightarrow \mathbb{R}_{\ge 0}$ and an initial ranking $\pi: V \rightarrow \mathbb{Z}$, this method proceeds in three main stages:
\begin{itemize}
\item Compute all strongly connected components ${S_1, S_2, \dots, S_k}$ in $G$.
\item Construct a DAG $D$ where each node corresponds to an SCC and directed edges represent cross-component links.
\item Topologically sort the DAG, and for each SCC, determine an internal node order to maximize forward edge weight.
\end{itemize}
For small SCCs (size $\leq 9$), the algorithm uses exhaustive permutation search to find the best internal node order. For large SCCs, it preserves the relative order from the input ranking $\pi$. The final output is a globally consistent node ranking $\pi'$ that respects the topological constraints between SCCs.

\subsubsection{Pseudocode}
\begin{algorithm}[H]
\caption{SCCBasedGlobalRanking}
\begin{algorithmic}[1]
\State \textbf{Input:} Directed graph $G = (V, E)$ with weights $w$, initial ranking $\pi$
\State Compute SCCs ${S_1, S_2, \dots, S_k}$ of $G$
\State Initialize $D \gets$ DAG with one node per SCC
\For{each edge $(u, v) \in E$}
\State Let $S_u, S_v$ be the SCCs containing $u$ and $v$
\If{$S_u \neq S_v$}
\State Add edge $(S_u, S_v)$ to $D$
\EndIf
\EndFor
\State Compute topological order $\sigma = [S_{i_1}, S_{i_2}, \dots, S_{i_k}]$ of DAG $D$
\State Initialize new ranking $\pi' \gets$ empty map, rank counter $\rho \gets 0$
\For{each SCC $S \in \sigma$}
\If{$|S| \leq 9$}
\State Try all permutations of $S$, let $O_S$ be ordering with max internal forward weight
\Else
\State Let $O_S \gets$ nodes in $S$ sorted by $\pi$
\EndIf
\For{each $v \in O_S$}
\State Set $\pi'(v) \gets \rho$; $\rho \gets \rho + 1$
\EndFor
\EndFor
\State Compute $\text{FW}{\text{new}} = \sum{(u,v) \in E,, \pi'(u) < \pi'(v)} w(u,v)$
\State Save $\pi'$ if $\text{FW}_{\text{new}}$ improves existing forward weight
\end{algorithmic}
\end{algorithm}

\subsubsection{Discussion}
This algorithm takes advantage of the acyclic structure of the SCC DAG to construct a globally consistent node ranking that aligns with the graph's natural directionality. By using exhaustive permutation only for small SCCs and defaulting to score-based sorting for larger components, the method balances optimality and efficiency. The global consistency of the final ranking ensures that backward edges across SCCs are minimized, making this approach well-suited for large graphs with hierarchical or modular structures.
If the resulting forward edge weight exceeds that of the input ranking, the new ranking is saved and used as an improved solution.

\section{Experiments}

\subsection{Dataset Description}

We evaluate our algorithms on the FlyWire Connectome Challenge dataset, a large-scale directed graph representing the neural connections in a Drosophila (fruit fly) brain. In this connectome, each node corresponds to a neuron and each weighted edge indicates the presence and strength of a synaptic connection from one neuron to another. The objective is to compute a total ordering of neurons that maximizes the sum of forward-pointing edge weights, thereby revealing an optimized feedforward flow of neural information.

The FlyWire connectome graph has the following properties:
\begin{itemize}
    \item \textbf{Number of nodes:} 136{,}648 (each node represents a neuron).
    \item \textbf{Number of edges:} 5{,}657{,}719 (each edge is a directed synapse with an associated weight).
    \item \textbf{Average degree:} 82.81 total connections per node (41.40 incoming edges on average, 41.40 outgoing edges on average).
    \item \textbf{Maximum degree:} 16{,}784 (the highest total number of in + out connections for a single neuron).
    \item \textbf{Median degree:} 54 (half of the neurons have degree less than or equal to 54).
    \item \textbf{Graph density:} 0.000303 (the ratio of existing edges to all possible directed edges, i.e.\ $\frac{|E|}{|V|(|V|-1)}$).
    \item \textbf{Weakly connected components (WCCs):} 160\,—\,subgraphs in which every neuron is reachable from every other when ignoring edge direction.
    \item \textbf{Strongly connected components (SCCs):} 9{,}626\,—\,maximal subgraphs in which each neuron can reach every other following edge direction.
    \item \textbf{Average SCC size:} 14.20 nodes (mean number of neurons per strongly connected component).
    \item \textbf{Largest SCC:} 126{,}840 nodes (the size of the biggest strongly connected component).
    \item \textbf{Singleton SCCs:} 9{,}503\,—\,SCCs of size one, i.e.\ neurons with no directed cycles involving other nodes.
    \item \textbf{Edge weights:} range from 2.0 to 2{,}405.0.
    \item \textbf{Mean edge weight:} 7.41, with a standard deviation of 12.77.
\end{itemize}

Due to the size and complexity of the graph, the diameter of the largest weakly connected component could not be computed. The structure of the connectome contains significant cyclic and modular substructures, making it a suitable benchmark for evaluating algorithms that aim to optimize forward edge weight.

This dataset was provided as part of an open optimization challenge hosted by Princeton University, where participants were tasked with producing a node ordering that reveals the underlying information flow structure in the brain. The challenge emphasizes algorithmic innovation, and participants are encouraged to improve upon a provided baseline by developing their own scoring and ranking strategies.

\subsection{Comparison with Prior Leading Solution}

To assess the effectiveness of our method, we compare our best node ranking against the strongest publicly available solution at the time of writing, implemented by David A. Bader, Harinarayan Asoori Sriram, Srijith Chinthalapudi, and Zhihui Du (denoted hereafter as \textbf{BA-CD}). Both solutions were evaluated on the full FlyWire connectome graph.

The key metrics are summarized below:

\begin{itemize}
    \item \textbf{Number of backward edges:} Our method resulted in 1,156,812 backward edges, compared to 1,157,173 in BA-CD.
    \item \textbf{Total backward edge weight:} Ours = 6,449,216.00; BA-CD = 6,452,875.00.
    \item \textbf{Forward edge weight (objective value):}
        \begin{itemize}
            \item BA-CD: $35,\!459,\!266 / 41,\!912,\!141 \approx 0.8460$
            \item Ours: $35,\!462,\!925 / 41,\!912,\!141 \approx 0.8461$
        \end{itemize}
    \item \textbf{Back edge length (node distance) statistics:}
        \begin{itemize}
            \item \textbf{BA-CD:} min = 1, max = 136{,}410, avg = 20{,}449.85, std = 24{,}081.19
            \item \textbf{Ours:} min = 1, max = 126{,}650, avg = 20{,}536.34, std = 24{,}542.31
        \end{itemize}
    \item \textbf{Back edge weight statistics:}
        \begin{itemize}
            \item \textbf{BA-CD:} avg = 5.58, std = 8.91
            \item \textbf{Ours:} avg = 5.57, std = 8.90
        \end{itemize}
\end{itemize}

Our method improves the total forward edge weight by 3,659 units over BA-CD, corresponding to an increase of approximately 0.0087\% in the total edge weight. This marginal but meaningful gain demonstrates our method's ability to more precisely reduce cyclic dependencies and enhance the feedforward structure in the neural graph.

\begin{figure}[H]
\centering
\includegraphics[width=0.95\textwidth]{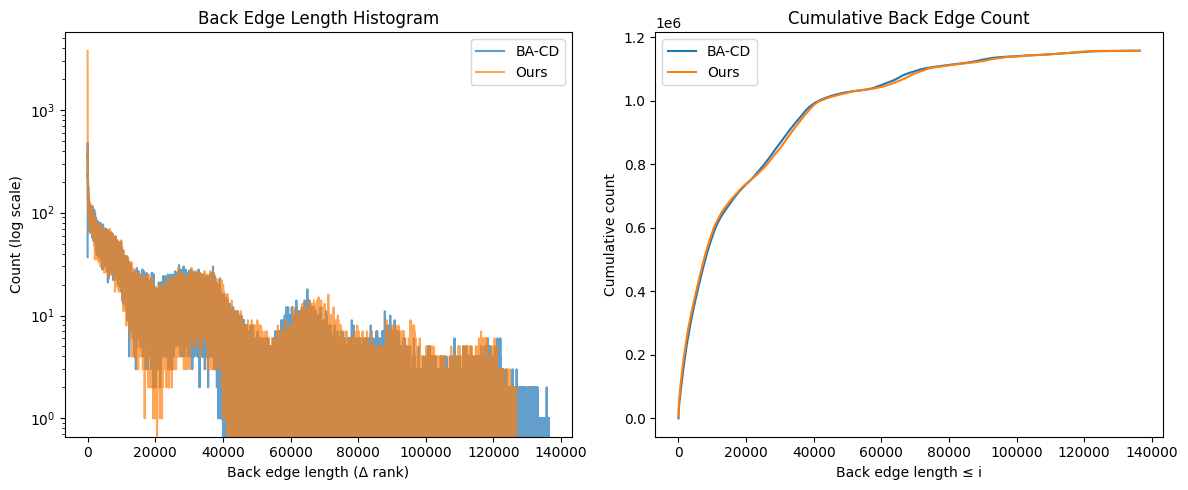}
\caption{Back edge length histogram (log scale, left) and cumulative back edge count (right) comparing our method and the BA-CD solution.}
\label{fig:back_edge_dist}
\end{figure}

Figure~\ref{fig:back_edge_dist} provides further insight into the structure of back edges. The histogram (left) shows that both solutions yield a heavy-tailed distribution, with most backward edges being short and a long tail of large-rank violations. Our method slightly reduces the frequency of long back edges over 120,000 ranks, which is consistent with the observed decrease in maximum back edge length. The cumulative plot (right) confirms that our method leads to a marginally earlier saturation in the total count of backward edges, suggesting that our ranking reduces the number of long-range feedback arcs that are particularly undesirable in modeling feedforward information flow.

\paragraph{Experimental Environment.} All experiments were conducted using Google Colab Pro+ with the High-RAM CPU runtime. This environment provides up to 32 GB of RAM and high-performance Intel Xeon CPUs with long-running session capabilities, enabling us to process the 136,648-node, 5.6-million-edge FlyWire connectome efficiently. The cloud-based environment allowed for fast prototyping and full-scale execution without relying on local hardware. All code was implemented in Python using standard scientific libraries such as NumPy, Pandas, and NetworkX.\\

\noindent
Table~\ref{tab:leaderboard} shows the leaderboard for the FlyWire Connectome Challenge as of June 2, 2025. Xin Zheng, Runxuan Tang, and Tatsuo Okubo implement three steps: (1) sort neurons by \(\mathrm{out\_degree}/\mathrm{in\_degree}\), yielding 30,136,267 (71.9\%) on FlyWire via \texttt{Beckers\_heuristic.ipynb}; (2) apply the graph reduction + divide-and-conquer (RASstar) to reach 33,037,112 (78.8 \%) \cite{Z24}; and (3) refine any ordering (e.g., RASstar’s) using the insertion-swap ILS method of \cite{S04}. The full information of their method and their codes can be found at \url{https://github.com/CIBR-Okubo-Lab/min_feedback_final}.
The solution by Ellis-Joyce, Both, and Turaga was declared the winner during the official challenge period. Based on the formal Github repository of their solution at \url{https://github.com/TuragaLab/connectome_reordering}, their approach was highly inspired by \cite{B24}. Higher-ranking submissions were made after the challenge deadline. The full leaderboard is available at: \url{https://codex.flywire.ai/app/mfas_challenge?dataset=fafb}. 

\vspace{1em}

\begin{table}[H]
\centering
\begin{tabular}{|l|p{7cm}|r|r|}
\hline
\textbf{Date} & \textbf{Name(s)} & \textbf{Forward Weight Sum} & \textbf{Percentage} \\
\hline
2025-06-02 & Soroush Vahidi, Ioannis Koutis & 35,462,925 & 85\% \\
2024-11-16 & David A. Bader, Harinarayan Asoori Sriram, Srijith Chinthalapudi, Zhihui Du & 35,459,266 & 85\% \\
2024-10-28 & Dritan Hashorva & 35,452,425 & 85\% \\
2024-10-28 & Justin Ellis-Joyce, Gert-Jan Both, Srinivas Turaga & 35,436,406 & 85\% \\
2024-10-08 & Justin Ellis-Joyce, Gert-Jan Both, Srinivas Turaga* & 35,435,948 & 85\% \\
2024-10-08 & Xin Zheng, Runxuan Tang, Tatsuo Okubo & 35,374,656 & 84\% \\
2024-10-08 & Dritan Hashorva & 35,364,447 & 84\% \\
2024-10-08 & David A. Bader, Harinarayan Asoori Sriram, Srijith Chinthalapudi, Zhihui Du & 35,231,406 & 84\% \\
2024-10-08 & Jose Estevez & 32,660,468 & 78\% \\
2024-10-07 & Thiago Munhoz da Nóbrega & 31,584,174 & 75\% \\
2024-10-08 & Rahul Gupta & 29,353,623 & 70\% \\
2024-09-24 & Dustin Garner & 29,027,621 & 69\% \\
2024-06-13 & Benchmark & 29,023,882 & 69\% \\
\hline
\end{tabular}
\label{tab:leaderboard}
\end{table}

The source code for all of our algorithms and experiments is publicly available at:
\url{https://github.com/SoroushVahidi/minimum-feedback-challenge}.

\section{Conclusion and Future Work}

We presented a suite of algorithms for computing node orderings that minimize the total weight of feedback arcs in large-scale directed graphs, with the goal of uncovering meaningful feedforward structures in neural connectomes. Our methods combine greedy scoring, adaptive local refinement, and global analysis of strongly connected components to optimize the forward edge weight in the FlyWire Connectome Challenge dataset.

Experimental results show that our best solution slightly outperforms the previously top-performing ranking in terms of forward edge weight, while preserving biological plausibility. We also analyze the structure of backward edges and provide insights into their distribution and magnitude. All algorithms are implemented in Python and were executed efficiently on Google Colab Pro+ with high-RAM CPU settings, supporting reproducibility without the need for high-performance computing infrastructure.

This work lays the foundation for several future directions. One promising avenue is to incorporate learning-based models to guide node ranking using structural or biological features. Extending the framework to dynamic or time-varying graphs may help model plasticity in neural circuits. Further, developing parallel versions of refinement steps could accelerate computation on massive datasets. From a theoretical standpoint, exploring approximation bounds or optimality guarantees in structured subgraphs could deepen our understanding of the feedback arc minimization problem in real-world networks.

We hope that the methods and insights presented here contribute both to algorithmic research and to the analysis of brain connectomics. We paraphrased several sentences with the assistance of ChatGPT, a language model developed by OpenAI \cite{ChatGPT}.

\bibliographystyle{plain}
\bibliography{ref}

\end{document}